\documentclass{article}

 \usepackage[preprint]{neurips_2026}

\usepackage[utf8]{inputenc} 
\usepackage[T1]{fontenc}    
\usepackage{hyperref}       
\usepackage{url}            
\usepackage{booktabs}       
\usepackage{amsfonts}       
\usepackage{nicefrac}       
\usepackage{microtype}      
\usepackage{xcolor}         
\usepackage{graphicx} 
\usepackage{amsmath}
\title{The Geometric Price of Discrete Logic: Context-driven Manifold Dynamics of Number Representations}

\author{%
  Long Zhang\thanks{Corresponding authors: longzhang@scut.edu.cn, cschenwn@scut.edu.cn} \quad 
  Dai-jun Lin\quad 
  Wei-neng Chen\footnotemark[1] \\
  School of Computer Science and Engineering \\
  South China University of Technology \\
  Guangzhou City, Guangdong Province, China \\
  \texttt{\{longzhang, cschenwn\}@scut.edu.cn} \\
}

\begin{document}

\maketitle

\begin{abstract}
Large language models (LLMs) generalize smoothly across continuous semantic spaces, yet strict logical reasoning demands the formation of discrete decision boundaries. Prevailing theories relying on linear isometric projections fail to resolve this fundamental tension. In this work, we argue that task context operates as a non-isometric dynamical operator that enforces a necessary "topological distortion." By applying Gram-Schmidt decomposition to residual-stream activations , we reveal a dual-modulation mechanism driving this process: a class-agnostic topological preservation that anchors global structure to prevent semantic collapse, and a specific algebraic divergence that directionally tears apart cross-class concepts to forge logical boundaries. We validate this geometric evolution across a gradient of tasks, from simple mapping to complex primality testing. Crucially, targeted specific vector ablation establishes a strict causal binding between this topology and model function: algebraically erasing the divergence component collapses parity classification accuracy from 100\% to chance levels (38.57\%). Furthermore, we uncover a three-phase layer-wise geometric dynamic and demonstrate that under social pressure prompts, models fail to generate sufficient divergence. This results in a "manifold entanglement" that geometrically explains sycophancy and hallucination. Ultimately, our findings revise the linear-isometric presumption, demonstrating that the emergence of discrete logic in LLMs is purchased at an irreducible cost of topological deformation.
\end{abstract}

\vspace{1em}

\section{Introduction}

The broad generalization capabilities of large language models (LLMs) are largely attributed to their internal continuous and smooth semantic representational topologies \citep{park_linear_2023}. However, executing logical tasks such as mathematical reasoning or strict classification requires the formation of decision boundaries within the feature space \citep{hu_representational_2026}. A fundamental tension exists between the continuity of semantic topology and the discrete nature of logical computation. While a continuous space allows for smooth transitions and similarity-based generalization between concepts, precise logical discrimination demands a black-and-white segregation of specific concepts. Therefore, how models traverse a shared continuous semantic space to dynamically forge discrete logical boundaries under specific task contexts constitutes a core challenge in understanding the internal mechanisms of LLMs.

At the Transformer architecture level, task instructions and contextual information modulate internal representations primarily via the residual stream. When processing context, attention mechanisms inject specific interference vectors into the residual stream of target concepts \citep{yang_unifying_2025}. Acting as a steering signal, these contextual interference vectors push the basal semantic representations toward specific task subspaces \citep[cf.][]{xu_low-dimensional_2026}. Interrogating the geometric properties and algebraic signatures of these interference vectors is an essential pathway to unraveling how models adapt to complex logical constraints.

Current research in representation geometry and mechanistic interpretability typically frames this contextual modulation through the lens of ``linear transformations.'' Studies employing linear probing and orthogonal subspace projections observe that task contexts tend to rigidly map conceptual relations into new orthogonal subspaces \citep{hu_representational_2026}. Under this linear representation hypothesis, contextual interference is generally treated as an orthogonal translation vector; researchers presume that the underlying geometric distances and topological structures of concepts remain highly consistent before and after the transformation, a state of isometric isomorphism. This perspective provides a mathematically tractable framework for explaining model translations across simple semantics or styles.

While the linear isometric hypothesis elegantly explains simple tasks, it falls fundamentally short in capturing the representational dynamics inherent in complex logical reasoning. Strict logical classification requires the model to forcibly tear apart concepts that are semantically highly similar but belong to different categories under specific logical rules \citep{hu_representational_2026}. Purely linear orthogonal projections cannot selectively compress or stretch the geometric distances between specific concept pairs without disrupting the global manifold \citep{zhou_geometry_2025}. Consequently, a glaring mechanistic gap remains: when continuous basal semantics geometrically misalign with the non-linear structural demands of discrete logical boundaries, by what mechanism does the model resolve this topological impediment?

This limitation motivates the core question of our study: how does the task context dynamically reshape the underlying concept manifold to satisfy discrete logical classification boundaries while simultaneously preventing the catastrophic collapse of basal semantic structures? If this representational transformation is not a simple linear orthogonal projection, it is imperative to precisely define the algebraic nature of contextual interference vectors and quantify their impact on the topological deformation of the representation space.

To bridge this conceptual chasm, we propose that contextual instructions are not merely translation vectors in coordinate space; rather, they act as modulation signals that trigger non-isometric manifold deformation. We argue that the specific perturbations introduced by the task context into the residual stream exert a dual modulation effect on the concept manifold. On one hand, a topological preservation mechanism ensures that similar basal concepts generate high covariance during network forward propagation, safeguarding against semantic collapse when steering toward the task subspace \citep[cf.][]{hu_representational_2026}. On the other hand, to satisfy specific logical classification boundaries, non-isometric manifold deformation injects directionally divergent pure innovation components between cross-class categories. This class-specific divergence shatters the original geometric metric, forcing the manifold to undergo localized topological deformation. Under this framework, geometric distortions in high-dimensional space are not negligible decoupling noise, but rather the irreducible algebraic cost the model must pay to forcefully transcend continuous semantics and execute discrete logic.

Empirically, we constructed an experimental paradigm based on algebraic decomposition and causal intervention. By employing Gram-Schmidt orthogonal decomposition, we stripped away the global translation vector to precisely isolate the pure specific interference vectors, establishing an ideal isometric isomorphic line as the absolute algebraic baseline to quantify topological compression and tearing. Furthermore, we designed a specific vector ablation experiment during forward inference. By algebraically erasing the specific topological deformation components in real-time within the computational graph, we directly observed changes in the model's logical classification function under a state of pure geometric isomorphism, thereby establishing a strict causal binding between topological deformation and the generation of logical boundaries.

By analyzing a gradient of tasks ranging from simple isomorphism to complex logic, we confirm that the realization of logical classification strictly relies on specific topological deformations. Crucially, ablation experiments demonstrate that wiping out this non-isometric distortion causes severe manifold entanglement between same-class and cross-class concepts, triggering a precipitous collapse in logical reasoning capabilities. Additionally, geometric analysis of sycophantic outputs reveals that when socially-induced interference lacks the specific divergence required to separate classes, the manifold fails to forge effective logical boundaries, providing a foundational geometric explanation for the model's blind compliance. These findings revise the theoretical presumption of absolute linear transformations in representation geometry, demonstrating that the emergence of intelligent behavior in LLMs is the product of a precise micro-level antagonism between topological preservation and specific deformation.
\section{Theoretical Framework \& Hypotheses}

\subsection{Base State: Initial Spatial Mapping of Concepts}
Let $x_i, x_j \in \mathbb{R}^d$ be the residual stream vectors at a given hidden layer under the baseline task (i.e., devoid of specific logical constraints), with their normalized directional vectors denoted as $\hat{x}_i = x_i / \|x_i\|$. The base similarity ($S_{base}$) between concepts represents the model's initial continuous semantic topology, defined as:
$$S_{base} = \langle \hat{x}_i, \hat{x}_j \rangle$$

\subsection{Vector Decomposition of Task Instructions (Gram-Schmidt Orthogonalization)}
Upon the introduction of a logical task instruction, the context induces a deformation of the underlying manifold, applying an interference vector $\Delta_i$ to the original concept. To decouple the dual effects of this operator, we employ Gram-Schmidt orthogonalization to precisely decompose $\Delta_i$ into a ``collinear component'' that anchors the basal semantics, and an ``orthogonal innovation component'' that drives functional differentiation:
$$\Delta_i = \|\Delta_i\| (\cos(\phi_i)\hat{x}_i + \sin(\phi_i)\hat{u}_i)$$
Here, $\cos(\phi_i)$ represents the projection ratio of the interference force along the original concept's direction, dictating the anchoring strength of the basal semantics; $\hat{u}_i$ denotes the pure innovation direction (a novel functional push catalyzed by the contextual instruction), which satisfies strict local orthogonality with the original concept, i.e., $\langle \hat{x}_i, \hat{u}_i \rangle = 0$.

\subsection{State Updates and Equivalent Rotation under RMSNorm}
The updated state is $x_i' = x_i + \Delta_i$. Defining the relative conflict intensity as $\omega_i = \|\Delta_i\| / \|x_i\|$, we expand the equation:
$$x_i' = \|x_i\| [(1 + \omega_i \cos(\phi_i))\hat{x}_i + \omega_i \sin(\phi_i)\hat{u}_i]$$
Taking RMSNorm, widely adopted in LLMs, as an example \citep{zhang_root_2019}, it incorporates root-mean-square scaling and learnable affine transformation weights $g$ (i.e., $y = x / \text{RMS}(x) \odot g$). This confines the output state to a high-dimensional ellipsoid in standard Euclidean space. To rigorously preserve angle-based geometric equivalence, we introduce a weighted inner product metric induced by the affine weights $g$ (defined as $\langle a, b \rangle_G = \sum g_k^2 a_k b_k$). Under this metric perspective, the high-dimensional ellipsoid is geometrically strictly equivalent to a perfect hypersphere. Since $\hat{x}_i \perp \hat{u}_i$, the norm scaling coefficient $N_i$ for the new state is:
$$N_i = \sqrt{1 + 2\omega_i \cos(\phi_i) + \omega_i^2}$$
Following spherical normalization, vector translation in the original space is strictly mapped to a rotation operation on the equivalent hypersphere. The new state $\hat{x}_i' = x_i' / \|x_i'\|$ can be formulated as the original concept vector $\hat{x}_i$ rotated by an angle $\alpha_i$ towards the innovation direction $\hat{u}_i$:
$$\hat{x}_i' = \cos(\alpha_i)\hat{x}_i + \sin(\alpha_i)\hat{u}_i$$
where the equivalent rotation angle $\alpha_i$ is entirely determined by the interference intensity and angle:
$$\cos(\alpha_i) = \frac{1 + \omega_i \cos(\phi_i)}{N_i}, \quad \sin(\alpha_i) = \frac{\omega_i \sin(\phi_i)}{N_i}$$
This rotational nature dictates a profound geometric reality: the model cannot add new features without paying a geometric cost; any movement toward the task subspace necessarily entails the compression or stretching of the original topological structure.

\subsection{Core Similarity Evolution Formula and Algebraic Cost Deconstruction}
Following a task switch, the new similarity $S_{new}$ between two concepts in the task subspace is the inner product of their rotated vectors:
$$S_{new} = \langle \hat{x}_i', \hat{x}_j' \rangle = \langle \cos(\alpha_i)\hat{x}_i + \sin(\alpha_i)\hat{u}_i, \cos(\alpha_j)\hat{x}_j + \sin(\alpha_j)\hat{u}_j \rangle$$
Expanding this expression utilizing the bilinearity of the inner product yields the strict core equation of manifold evolution:
$$S_{new} = \cos(\alpha_i)\cos(\alpha_j)S_{base} + \cos(\alpha_i)\sin(\alpha_j)\langle \hat{x}_i, \hat{u}_j \rangle + \sin(\alpha_i)\cos(\alpha_j)\langle \hat{u}_i, \hat{x}_j \rangle + \sin(\alpha_i)\sin(\alpha_j)\langle \hat{u}_i, \hat{u}_j \rangle$$
This formula elegantly dissects topological deformation into three distinct mechanisms:
\begin{itemize}
    \item \textbf{Base Cosine Similarity ($S_{base}$):} The retention rate of the basal structure, governed by the rotation angle $\alpha$.
    \item \textbf{Topological Preservation ($C_{ij} = \langle \hat{x}_i, \hat{u}_j \rangle$):} Measures the extent to which the pure push ($\hat{u}_j$) applied to concept $j$ ripples into the original location of concept $i$ ($\hat{x}_i$).
    \item \textbf{Specific Divergence ($U_{sim} = \langle \hat{u}_i, \hat{u}_j \rangle$):} Captures the directional synergy or divergence of the specific pushes applied to different concepts to accomplish the task (e.g., the directional discrepancy between $\hat{u}_i$ applied to $i$ and $\hat{u}_j$ applied to $j$).
\end{itemize}

To parse the algebraic properties of specific divergence $U_{sim}$, let $v_i$ be the specific interference vector applied to sample $i$. To isolate the pure innovation direction $\hat{u}_i$, we project $v_i$ onto the normal plane of $\hat{x}_i$ and normalize it. Let $p_i = \langle v_i, \hat{x}_i \rangle$ be the projection length of the interference along the original direction, and $q_i = \|v_i - p_i \hat{x}_i\|$ be the length of the orthogonal component. The pure innovation direction is exactly $\hat{u}_i = (v_i - p_i \hat{x}_i) / q_i$. Substituting this into $U_{sim} = \langle \hat{u}_i, \hat{u}_j \rangle$ and utilizing $\langle \hat{x}_i, \hat{x}_j \rangle = S_{base}$, we obtain:

\begin{align*}
U_{sim} &= \left\langle \frac{v_i - p_i \hat{x}_i}{q_i}, \frac{v_j - p_j \hat{x}_j}{q_j} \right\rangle \\
&= \frac{1}{q_i q_j} (\langle v_i, v_j \rangle - p_j \langle v_i, \hat{x}_j \rangle - p_i \langle \hat{x}_i, v_j \rangle + p_i p_j \langle \hat{x}_i, \hat{x}_j \rangle) \\
&= \frac{p_i p_j}{q_i q_j} S_{base} + \frac{\langle v_i, v_j \rangle - p_j \langle v_i, \hat{x}_j \rangle - p_i \langle \hat{x}_i, v_j \rangle}{q_i q_j}
\end{align*}

By defining the slope $\lambda = (p_i p_j) / (q_i q_j)$ and the intercept term $k = (\langle v_i, v_j \rangle - p_j \langle v_i, \hat{x}_j \rangle - p_i \langle \hat{x}_i, v_j \rangle) / (q_i q_j)$, we derive the linear trend equation:
$$U_{sim} = \lambda S_{base} + k$$
This mathematically proves that the innovation direction is not random noise, but exhibits a profound linear coupling with $S_{base}$.

\subsection{Core Theoretical Hypotheses}
In summary, the modulation of basal concept representations by task context is not a simple Linear Orthogonal Projection, but a context-driven non-isometric manifold deformation. The specific perturbation ($\Delta_{specific}$) injected into the residual stream exerts a dual modulation effect:
\begin{itemize}
    \item \textbf{Topological Preservation (via $C$):} High covariance among similar basal concepts during forward propagation induces cross-linkages in the perturbation vectors ($C_{ij} > 0$). This preservation ensures the manifold averts catastrophic semantic collapse when steered toward the task space.
    \item \textbf{Non-isometric Manifold Deformation (via $U$):} To forge specific logical classification boundaries, the perturbation operator injects directionally divergent pure innovation components ($U_{sim} \leq 0$) between particular categories. This Specific Divergence shatters the original geometric metric, compelling the manifold to undergo localized topological deformation to form linearly separable decision clusters.
\end{itemize}

Based on this framework, we propose five core hypotheses (H1-5):
\begin{itemize}
    \item \textbf{H1 (Law of Topological Preservation):} During steering toward the task subspace, interdependencies exist within the basal representation network ($C_{ij} > 0$). This topological preservation resists drastic structural alterations, preventing semantic collapse.
    \item \textbf{H2 (Isomorphic Translation Baseline):} The higher the base similarity between concepts, the more congruent their task interference directions, yielding a significant positive linear correlation between $U_{sim}$ and $S_{base}$.
    \item \textbf{H3 (Specific Deformation Gradient):} In logical classification tasks, the intercept term $k_{cross}$ for cross-class concept pairs is lower than $k_{same}$ for same-class pairs, indicating that same-class concepts receive more aligned task interference.
    \item \textbf{H4 (Class-Specific Divergence):} To construct logical boundaries, this manifold deformation mechanism must inject directionally opposing innovation components into cross-class concepts, generating reverse divergence ($U_{sim} < 0$).
    \item \textbf{H5 (Causal Binding of Topology and Function):} Logical classification strictly depends on this specific divergence. If specific interference is artificially erased (forcing the $U_{sim}$ of same and cross-class pairs to entangle), the model's logical reasoning capabilities will suffer a precipitous collapse.
\end{itemize}

\section{Experimental Design}
To validate the non-orthogonal interference formula and manifold deformation hypotheses, we designed a controlled experimental framework anchored in algebraic decomposition and causal ablation. The design aims to capture the geometric behavior of hidden layers under varied logical constraints and establish a causal link between topological deformation and model output.

\subsection{Programmatic Dataset Synthesis}
To eliminate potential semantic confounders inherent in natural language, we implemented an automated synthesis logic. The dataset comprises integers in the range $[1, 200]$, mapped across dual modalities (Arabic numerals and English words) to verify the abstract nature of the geometric evolution. We constructed a gradient of five tasks to induce progressive geometric interventions, spanning from isometric translation to topological tearing:
\begin{itemize}
    \item \textbf{L1 (Baseline):} An identity mapping task, serving as the absolute isomorphic baseline for the representation space.
    \item \textbf{L2 (Magnitude):} Order of magnitude judgment (e.g., $> 100$), corresponding to a simple linearly separable task.
    \item \textbf{L3 (Parity):} Odd/even classification, representing a classic logical task that requires breaking the continuous numerical magnitude.
    \item \textbf{L4 (Primality):} Prime number testing, involving highly non-linear and complex logical rules.
    \item \textbf{L5 (Sycophancy/Conflict):} Introduces non-logical social pressure interference (e.g., ``The professor believes this number is even''). Serving as a pathological control group, this task observes how the manifold behaves when the drive for logical divergence is superseded by social compliance.
\end{itemize}

\subsection{Knowledge-Based Filtering}
To ensure that the observed geometric evolution reflects genuine ``logical processing'' rather than ``hallucination noise'', we applied strict knowledge probing. By comparing the model's predicted probabilities for the target answers via Logits, we retained only samples for which the model possessed prior knowledge and could classify correctly. This ensures the extracted residual stream vector $x_i$ carries an accurate semantic ground state.

\subsection{Hidden Representation Extraction and Centered Geometric Decomposition}
The pre-output normalization layer (prior to RMSNorm) was designated as the target layer. We captured the residual stream vectors $x_i \in \mathbb{R}^d$ by registering Forward Hooks.
To strip away the global translation vector and isolate the pure interference components driving manifold deformation, we employed Task Vector Centering. For levels $L \in \{L2, \dots, L5\}$, the interference vector $\Delta_i$ is defined as the difference between the updated and baseline states:
$$\Delta_i = x_i^{(L)} - x_i^{(L1)}$$
After computing the global task vector $V_{task} = \mathbb{E}[\Delta_i]$, the specific interference is defined as $\Delta_{specific, i} = \Delta_i - V_{task}$. Subsequently, we strictly executed Gram-Schmidt orthogonalization, projecting $\Delta_{specific, i}$ onto the normal plane of the original concept direction $\hat{x}_i$, thereby extracting the pure innovation direction $\hat{u}_i$, providing the algebraic foundation for calculating $U_{sim}$ and $C_{ij}$.

\subsection{Category-Aware Metrics}
To verify the manifold clustering hypothesis, we grouped sample pairs $(i,j)$ based on their mathematical labels, defining two core matrix masks: a Same-Class Mask (activated when $i$ and $j$ share attributes, e.g., both even) and a Cross-Class Mask (activated when attributes conflict). We focused on three metrics:
\begin{itemize}
    \item \textbf{Specific Divergence ($U_{sim}$):} Determines whether interference directions across samples are synchronized, serving as the direct observation of specific divergence.
    \item \textbf{Topological Preservation ($C_{ij}$):} Measures the extent to which the new task preserves the original representation structure.
    \item \textbf{Pearson Correlation ($r$):} Assesses the alignment between $S_{base}$ and $U_{sim}$, acting as the statistical criteria for deviations from isometric isomorphism.
\end{itemize}

\subsection{Manifold Healing via Specific Vector Ablation}
To establish the causal relationship between geometric tearing and logical behavior (H5), we designed two ``Specific Vector Ablation'' experiments. First, during forward inference, we utilized pre-hooks to perform real-time algebraic erasure of the specific interference vector responsible for manifold clustering (Direct Intervention):
$$x_{patched} = x_{original} - \Delta_{specific, label}$$
Second, let the interference vector for a specific category be $v_{label}$ (i.e., $\Delta_{specific, label}$). Instead of directly subtracting $v_{label}$ from the base state $x_{original}$, we projected $v_{label}$ onto the normal plane of $x_{original}$ during real-time inference, subtracting only the orthogonal component. Letting the normalized original direction be $\hat{x} = x_{original} / \|x_{original}\|$, the projection (collinear component) is $p = \langle v_{label}, \hat{x} \rangle \hat{x}$, and the orthogonal innovation component is $v^\perp = v_{label} - p$. This yields the second ablation formula (Ortho Intervention):
$$x_{patched\_new} = x_{original} - v^\perp$$
Both experiments were evaluated across two dimensions:
\begin{itemize}
    \item \textbf{Geometric Evaluation:} Observing whether the forced ``healing'' causes the $U_{sim}$ of same and cross classes to collapse back into an entangled state.
    \item \textbf{Functional Behavior Evaluation:} Observing whether the accuracy of specific classification tasks (e.g., parity) suffers significant degradation following the ablation.
\end{itemize}
By comparing manifold states and task accuracy pre- and post-intervention, this experiment delivers the ultimate causal evidence that ``logical classification dictates an algebraic cost.''

\subsection{Layer-wise Geometric Dynamics Tracking}
Although extending beyond the pre-RMSNorm layer targeted by the theoretical framework, tracking the evolution of $U_{sim}$ and $C_{ij}$ across depth profoundly deepens our understanding of the layer-wise dynamics of numerical representations. By probing all hidden layers of the LLM, we quantified how representation evolves across network depth for reasoning tasks of varying complexities \citep[e.g.,][]{jin_exploring_2025}. The core methodology employs Gram-Schmidt decomposition at each layer to partition the activation tensor into a shared ``basal structural direction'' and a task-specific ``orthogonal innovation component''. Based on this, we tracked the cosine similarity of the innovation components (i.e., $U_{sim}$) and the structural entanglement (i.e., $C_{ij}$) for both Same-Class and Cross-Class pairs, microscopically unraveling the geometric phase transitions during feature extraction, logical computation, and final vocabulary alignment.

\section{Results \& Discussion}

This section systematically validates the non-isometric manifold deformation hypothesis proposed in our theoretical framework by synthesizing the geometric evolution panoramas and causal ablation data. To precisely quantify topological deformation, we first delineate the absolute algebraic baselines in Figure 1 based on the core equation $U_{sim} = \lambda S_{base} + k$: the ideal isometric isomorphic line ($y=x$) represents task interference perfectly preserving the geometric distance proportions between basal concepts; the orthogonal independence line ($y=0$) represents an interference direction completely orthogonally decoupled in the local space. These two baselines strictly partition the representation space into an isometric translation zone (distributed along $y=x$), a topological expansion zone ($0 < y < x$), and a specific divergence zone ($y < 0$).

\subsection{Rigor of Mathematical Derivation and Abstractness of Concepts}

Before delving into manifold evolution, we empirically validated the mathematical rigor of the non-orthogonal interference formula. Experimental results demonstrate that across tasks of varying complexity, the local inner product $\langle\hat{x}_i, \hat{u}_i\rangle$ between the updated state and its orthogonal pure innovation component consistently remains at the magnitude of $10^{-7}$. This remarkably low computational residual definitively proves the algebraic robustness of Gram-Schmidt decomposition when processing high-dimensional residual streams. More crucially, when the model processes two radically different surface modalities, ``Arabic numerals'' and ``English words'' \citep[cf.][]{hu_representational_2026}, the extracted specific interference vectors $\Delta_{specific}$ exhibit a cross-modal similarity of 0.8210. This high degree of cross-modal consistency confirms that our object of observation is not the shallow token embeddings, but rather the precise geometric modulation executed on a highly abstract Concept Manifold.

\subsection{Isomorphic Translation vs. Logical Boundaries: Aligning with H1 and H2}

By centering the task vectors and introducing category label masks, we observed the bifurcating behavior of representation geometry during task switching (as shown in Table 1 and Figure 1). The observational data reveals that manifold evolution is a precision process co-driven by topological preservation and specific divergence.

\begin{table}[htbp]
  \caption{Quantitative Summary of Core Geometric Metrics Across Task Levels}
  \label{tab:geometric_metrics}
  \centering
  \resizebox{\textwidth}{!}{
  \begin{tabular}{llcccccc}
    \toprule
    \multicolumn{2}{c}{\textbf{Task Info}} & \multicolumn{2}{c}{\textbf{Pearson $r$}} & \multicolumn{2}{c}{\textbf{Mean $U_{sim}$}} & \multicolumn{2}{c}{\textbf{Mean $C_{ij}$}} \\
    \cmidrule(r){1-2} \cmidrule(lr){3-4} \cmidrule(lr){5-6} \cmidrule(l){7-8}
    Level & Type & Same & Cross & Same & Cross & Same & Cross \\
    \midrule
    L2 & Magnitude (Simple)   & 0.8359 & 0.6254 & 0.6447 & $-0.1977$ & 0.3136 & 0.4342 \\
    L3 & Parity (Medium)      & 0.6011 & 0.4121 & 0.5763 & $-0.2351$ & 0.3353 & 0.3190 \\
    L4 & Primality (Complex)  & 0.6962 & 0.5756 & 0.4090 & $-0.1202$ & 0.3913 & 0.3308 \\
    L5 & Sycophancy (Induced) & 0.7659 & 0.7580 & 0.3508 & 0.3190  & 0.4547 & 0.4574 \\
    \bottomrule
  \end{tabular}
  }
\end{table}

\begin{figure}[t] 
  \centering
  \includegraphics[width=\textwidth]{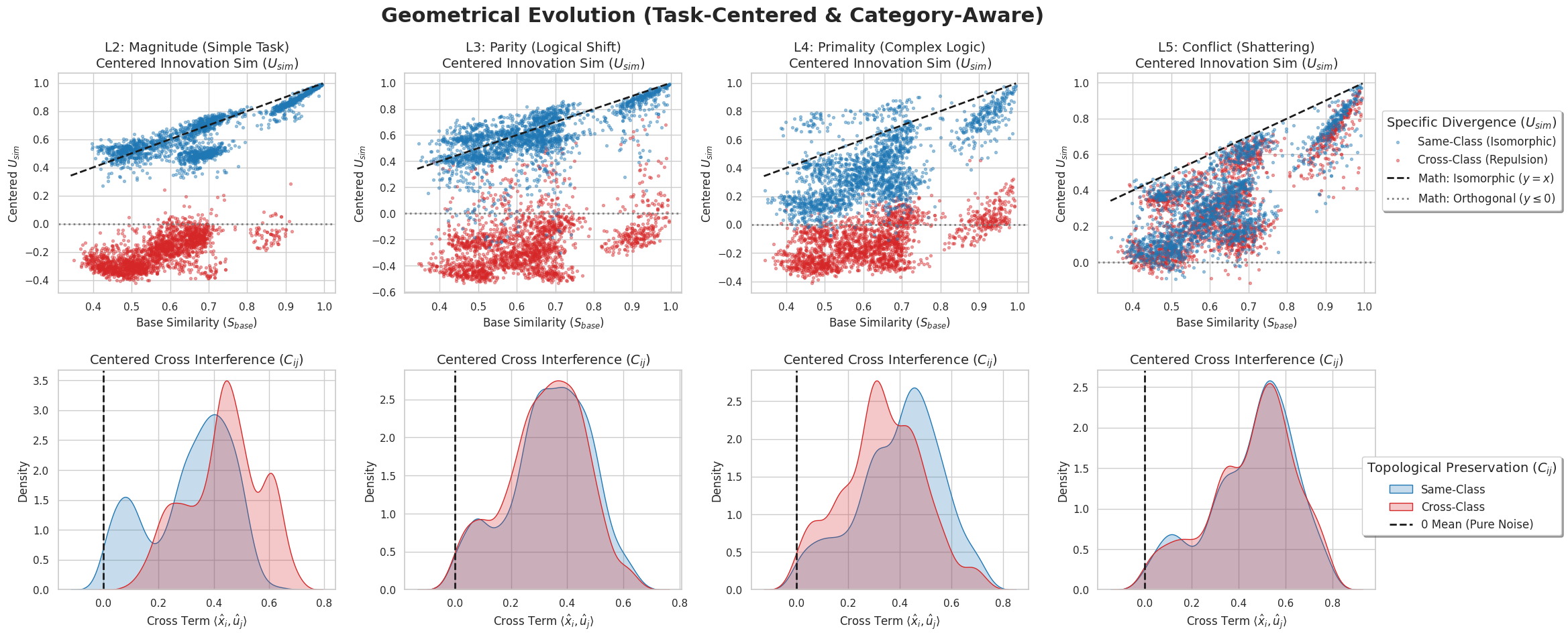} 
  \caption{Panoramic analysis of geometric evolution across task levels. The scatter plots illustrate the transition from isometric translation to specific divergence, while the bottom density plots confirm class-agnostic topological preservation ($C_{ij}$).}
  \label{fig:panoramic}
\end{figure}

First, regarding the maintenance mechanism of the underlying topology, the density plot of topological preservation ($C_{ij}$) at the bottom of Figure 1 provides a decisive finding. Across all task levels, the mean $C_{ij}$ not only stably remains positive (between 0.32 and 0.45), but the $C_{ij}$ distribution curves for Same-Class and Cross-Class samples nearly overlap (especially in L3, L4, and L5). This phenomenon profoundly validates Hypothesis 1 (Law of Topological Preservation). The data indicates that when introducing task interference, the model still applies an equivalent structural retention force to logically conflicting cross-class concepts as it does to same-class concepts. Rather than distinguishing concepts by severing their basal semantic connectivity, the network exhibits a uniform topological preservation, ensuring global stability when the manifold steers into the task subspace \citep[cf.][]{hu_representational_2026}.

Given the uniform $C_{ij}$ distribution, the generation of logical boundaries heavily relies on $U_{sim}$. In the simple task (L2: Magnitude), the scatter plot of same-class sample pairs (blue clusters) tightly adheres to the $y=x$ baseline (Pearson $r=0.8359$), strictly conforming to Hypothesis 2 (Isomorphic Translation Baseline). A similar positive correlation trend exists in L3, L4, and L5; however, the blue clusters in L3 are more dispersed, while those in L4 and L5 broadly enter the $y<x$ region. This indicates they do not entirely retain their proportional positions from the original numerical representation space, but undergo varying degrees of shape-diffusing distortion. Intriguingly, visual inspection reveals that cross-class samples (red clusters) maintain a structure similar to the blue clusters \citep[cf.][]{hu_representational_2026}, also positively correlated, but their positions have distinctly sunk. This validates Hypothesis 3 (Specific Deformation Gradient). While this is obvious in L2, L3, and L4, it is less apparent in L5. A plausible explanation is that the social pressure task in L5 is not a logical task and has little relevance to mathematical concepts \citep[cf.][]{zhou_lssf_2025}; thus, Qwen2.5 does not attempt to accurately distinguish numerical representations under this specific interference. In other words, the model has no intention of separating mathematical concepts here; hence, the interference directions remain similar for both same- and cross-class data.

We discovered that in the L2 task, the mean $U_{sim}$ for cross-class samples has already dropped to $-0.1977$, with the vast majority of the red cluster sinking into the specific divergence zone ($y<0$). This proves that even for a basic magnitude judgment task, the model must construct binary decision boundaries by injecting directionally opposing pure innovation components, directly validating Hypothesis 4 (Class-Specific Divergence). Observational data further shows that the mean cross-class $U_{sim}$ drops to $-0.2351$ and $-0.1202$ in L3 and L4 tasks, respectively, and the red clusters experience a pronounced subsidence, universally breaking below the $y=0$ orthogonal baseline. This demonstrates that to forcefully distinguish conflicting concepts, the network actively constructs adversarial interference directions, causing the original manifold to ``cluster'' \citep[cf. fig.3., p.6,][]{hu_representational_2026}.

Simultaneously, the scatter plots of same-class samples in L3 and L4 begin to deviate downward from the $y=x$ line. This reveals the global algebraic cost of logical computation: to forcibly separate continuous numerical concepts along specific dimensions, the model inevitably inflicts stretching and distortion upon the local topology of same-class concepts. In the L4 prime task, the cross-class $U_{sim}$ rebounds slightly ($-0.1202$) compared to L3, with a more fragmented distribution within the divergence zone. This counter-intuitive data actually reflects the geometric signature of complex logic: the distribution of primes within natural numbers is highly sparse and non-linear. The model struggles to find a unified, strong global orthogonal divergence direction for primes versus composites, unlike it does for parity \citep{hindupur_projecting_nodate, hu_representational_2026}. Consequently, manifold clustering in L4 exhibits a more fragmented topological deformation, weakening the overall reverse divergence mean. Combined with $C_{ij}$ observations, this establishes that the model executes an ``incremental logical clustering'' strategy: while maintaining basal topological preservation, it forcefully applies specific divergence ($U_{sim} < 0$) to carve out discrete decision clusters within the continuous space.

\subsection{Sycophancy Manifold Observation: Topological Collapse under Induced Interference}

In the L5 (Sycophancy/Conflict) task, we observed a geometric evolution pattern fundamentally distinct from logic-driven tasks. When non-logical social pressure interference is introduced, the model fails to trigger the expected class divergence mechanism. Table 1 displays an anomalous surge in the mean cross-class $U_{sim}$ to +0.3190. In the far-right scatter plot of Figure 1, the red clusters representing cross-class concepts fail to penetrate the $y=0$ line into the divergence zone; instead, they remain entirely trapped in the $0 < y < x$ topological expansion zone, suffering severe manifold entanglement with same-class concepts (blue clusters).

This phenomenon uncovers the essential geometric difference between logical conflict and social pressure in the representation space. Constructing divergence vectors ($U_{sim} < 0$) requires the model to pay a significant algebraic cost to overcome base similarity. The injection of strong external directives suppresses the model's ability to generate reverse divergence, causing the interference vectors to maintain positive synergy ($U_{sim} > 0$) across all concepts. Because cross-class concepts are not pushed into the opposing geometric quadrant, they become indistinguishable from same-class concepts within the task's innovation subspace. This failure to cross the $y=0$ boundary and form distinct logical clusters provides a foundational geometric explanation for hallucination and blind compliance in LLMs: the model fails to pay the requisite algebraic cost to shatter continuous semantics, resulting in the loss of decision boundaries and ultimate feature separation failure \citep{huang_survey_2025}. Here, the model's attention is likely hijacked by the necessity to comply with social pressure rather than focusing on the parity resolution task \citep[refer to Social-Information Competition Dynamics,][]{zhang_human-like_2026}.

\subsection{Causal Intervention: Functional Paralysis Induced by Specific Vector Ablation}

To definitively establish the strict causal link between geometric clustering and logical classification, we implemented specific vector ablation experiments based on algebraic erasure, validating Hypothesis 5 (Causal Binding of Topology and Function).

\begin{figure}[t]
  \centering
  \includegraphics[width=0.9\linewidth]{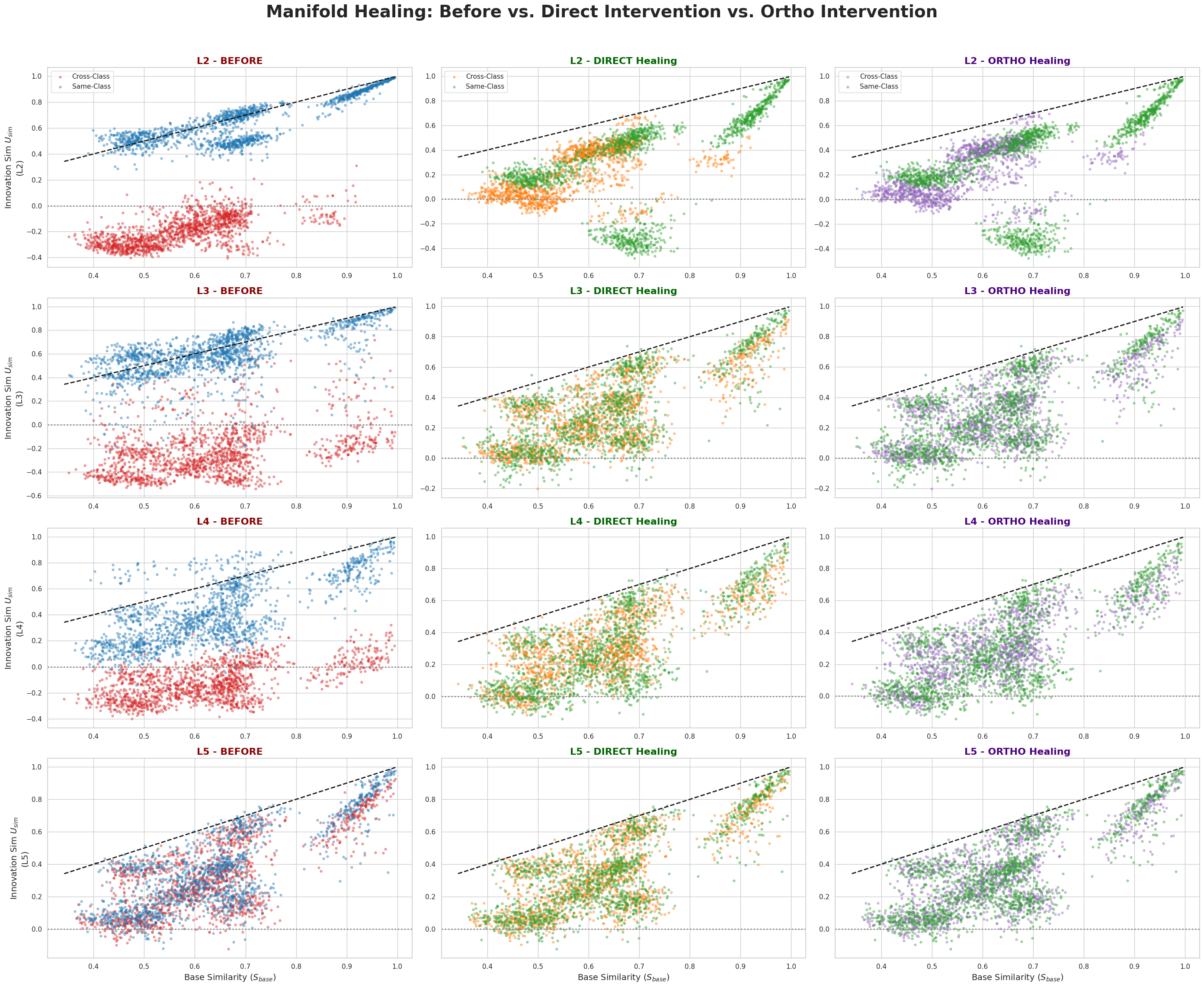}
  \caption{Comparison of specific vectors pre- and post-ablation. The specific vector erasure successfully collapses the cross-class clusters back into the entangled isomorphic zone.}
  \label{fig:ablation}
\end{figure}

As illustrated in Figure 2, regarding geometric evaluation, by performing real-time subtraction of the class-specific divergence vector $\Delta_{specific}$ and the orthogonal component $v^{\perp}$ during forward inference, we successfully reversed the topological morphology of the manifold. Visually, the cross-class clusters that originally sank into the $y<0$ region due to specific divergence were forcibly pulled back into the positive $0 < y < x$ zone, remixing with the same-class clusters. The visual and calculative similarities between the two healing methods are striking (similarity $> 0.98$). Taking L3 as an example, the ablation successfully wiped out the specific vectors, forcing the representation space to regress into a continuous entangled state devoid of logical boundaries.

Regarding functional behavior evaluation, this forced geometric healing was accompanied by a catastrophic collapse of the model's logical capabilities. In the L3 parity classification task, specific vector ablation caused the discrimination accuracy to plummet precipitously from 100.00\% to 38.57\%. This sub-random (<50\%) performance represents a severe adversarial functional paralysis. This profound degradation carries profound theoretical implications: once specific divergence is artificially stripped away, the model is forced to rely on the underlying continuous $S_{base}$ for discrimination. Because adjacent numbers (e.g., 2 and 3) are highly similar in the base space, this continuity creates a fatal misdirection for parity classification. This mirrors experiments in human cognition, where elements with similar semantics but different rule constraints increase error rates and cognitive load \citep[e.g.,][]{boot_eye_2022}.

It is crucial to emphasize that this performance plunge is not caused by injecting destructive adversarial noise into the hidden layers. The post-intervention manifold regression evaluation (Figure 2) clearly shows that the ablation did not cause the representation space to scatter randomly into a disordered state; rather, it with pinpoint accuracy forced the cross-class clusters to retrace their path back into the $0 < y < x$ isomorphic entanglement zone. This targeted topological closure confirms that our intervention is a highly specific algebraic inverse operation, not a random perturbation destroying the underlying distribution.

In summary, the ablation experiments provide compelling causal evidence: the topological clustering observed in non-isometric manifold deformation is absolutely not accompanying redundant noise; it is the indispensable algebraic cost the model must pay to traverse continuous semantics and forge discrete logical boundaries.

\subsection{Layer-wise Geometric Dynamics Tracking}

\begin{figure}[ht]
  \centering
  \includegraphics[width=\textwidth]{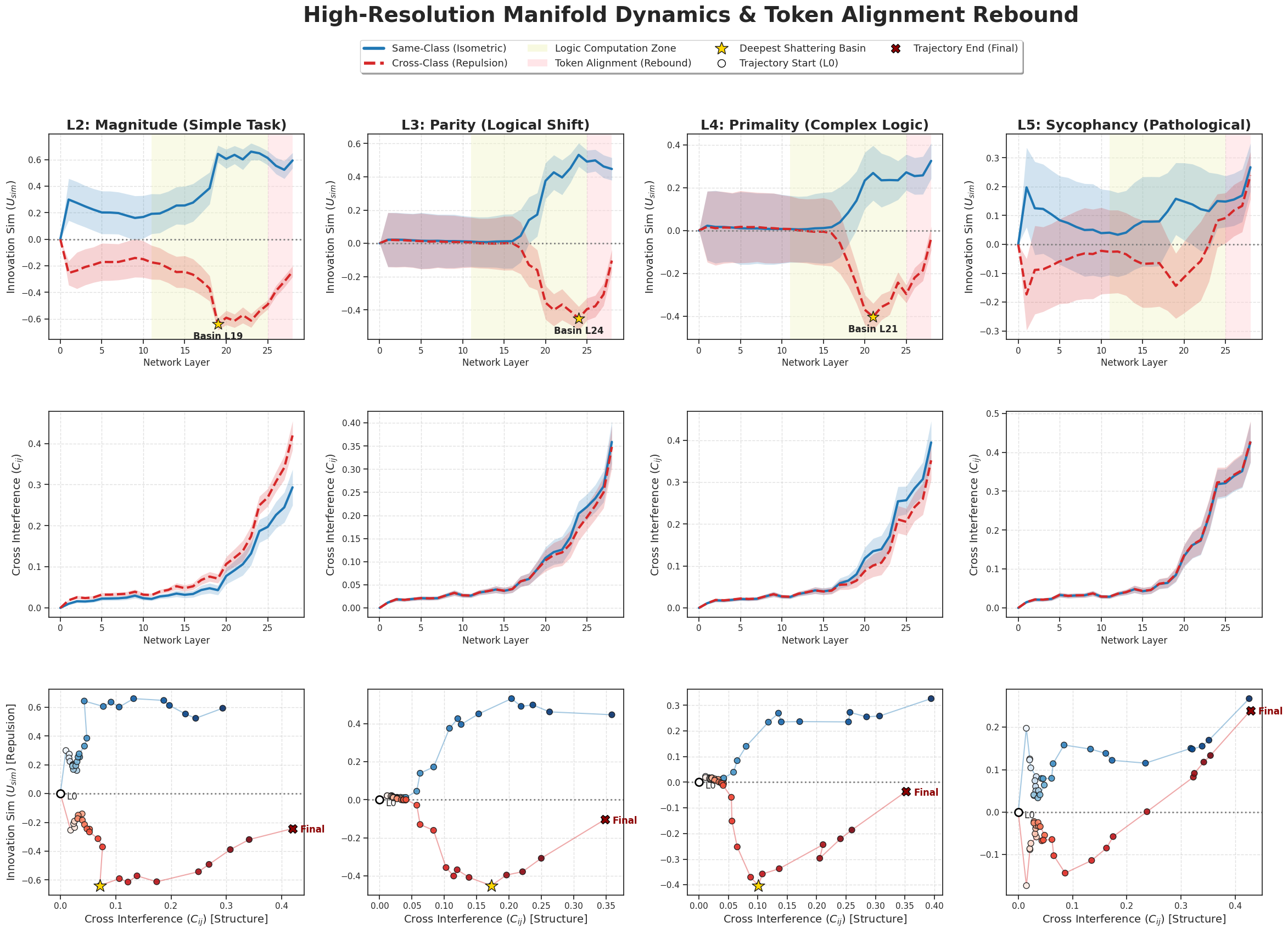}
  \caption{Layer-wise evolution of representation manifolds. Top row: $U_{sim}$ dynamics showing the three-phase mechanism; Middle row: $C_{ij}$ tracking; Bottom row: 2D phase portraits illustrating the scissor-like bifurcation in logical tasks.}
  \label{fig:layerwise}
\end{figure}

The first row of Figure 3 reveals a three-phase mechanism collectively followed by logical computation tasks across network depth, akin to decision-making processes \citep{joshi_geometry_2025}.

\textbf{Phase 1 (Shallow Extraction Zone):} The initial layers primarily perform semantic enrichment of token representations. Both same-class and cross-class $U_{sim}$ show no obvious differentiation, consolidating at low levels.

\textbf{Phase 2 (Deep Computation Basin):} Entering mid-to-deep layers, same-class $U_{sim}$ rises steadily, while cross-class $U_{sim}$ plunges sharply under the drive of specific divergence, forming a Minimum Cross-Class Similarity point at a specific layer. This geometric location, termed the ``basin'', is where the cost of generating logical boundaries is most concentrated.

\textbf{Phase 3 (Output-Layer Rebound Zone):} Nearing the output, cross-class $U_{sim}$ converges from the negative extreme back toward zero, exhibiting a unidirectional similarity rebound. Same-class $U_{sim}$ plateaus during this phase rather than rising significantly, highlighting asymmetric behavior between the two curves during rebound.

The layer location of the basin extreme varies by task logical structure and does not deepen monotonically with complexity. The L2 basin appears at layer 19, L3 is deepest at layer 24, while L4 regresses to layer 21, shallower than L3. This non-monotonic pattern perfectly aligns with the mechanism proposed in \S4.2: parity possesses a singular, globally consistent binary cut-plane, allowing the model to concentrate strong divergence vectors in deep layers to form a single deep basin. Conversely, primes are sparse and non-linear, making it difficult for the model to find a unified global orthogonal divergence direction; its geometric clustering is fragmented, resulting in a shallower basin with weaker extremes than the parity task.

Furthermore, L2 and L5 exhibit observable early-layer differentiation ($N<10$) in $U_{sim}$, whereas L3 and L4 show almost none until mid-deep layers. It must be stressed that the internal mechanisms behind these early differentiations are entirely different: L2's early differentiation stems from numerical magnitude being a surface heuristic easily captured by shallow attention; the numerical morphology of the input sequence carries sufficient signal. In contrast, any early variance in L5 reflects the shallow activation effects of external social pressure directives on attention weights, irrelevant to the logical computation of mathematical concepts. These should not be conflated as manifestations of the same mechanism.

The second row of Figure 3 plots the evolution of the topological preservation metric ($C_{ij}$). In logical tasks (L2 to L4), $C_{ij}$ curves for both classes show sustained positive accumulation and highly overlap across layers, reinforcing the finding that ``topological preservation is class-agnostic.'' Notably, the absolute value of $C_{ij}$ for L5 is globally elevated (consistent with Table 1) and its shallow accumulation slope is steeper, showing a discernible magnitude difference from logical tasks. This suggests that under social pressure, contextual interference vectors have a stronger structural entanglement with basal numerical semantics, the model fails to direct the interference toward logical separation, instead uniformly injecting global synergy across all concepts, further cementing the ``manifold entanglement'' interpretation in \S4.3.

The third row abstracts this evolution into a 2D phase portrait ($C_{ij}$ as X-axis, $U_{sim}$ as Y-axis), tracing a complete trajectory from L0 to Final. Same-class and cross-class trajectories exhibit opposing path structures. Same-class trajectories move primarily rightward ($C_{ij}$ accumulation) with a slight $U_{sim}$ rise in the computation zone, converging stably. Cross-class trajectories plummet vertically into negative territory during computation, hit the basin extreme, then are pulled rightward by continuous $C_{ij}$ accumulation, finally looping back near zero in the alignment zone. For logical tasks (L2 to L4), these trajectories form a right-opening ``scissor-like bifurcation'', with the bifurcation depth dictated by the $U_{sim}$ basin extreme.

Conversely, the L5 phase portrait shows marked pathological traits: the cross-class trajectory fails to enter the $U_{sim} < 0$ divergence zone. Both trajectories are highly entangled, failing to form any effective scissor-like bifurcation. From a phase-space perspective of manifold geometry, this provides supplementary evidence for the feature separation failure observed in \S4.3: social pressure tasks lack the specific divergence required to drive the cross-class trajectory downward, causing logical differentiation to completely fail at the phase-space level.

\section{Implications \& Conclusion}

\subsection{Implications \& Suggestions}

This study not only illuminates the reconciliation mechanism between continuous semantics and discrete logic but also provides a novel geometric perspective for Internal Alignment and architectural optimization of LLMs.

First, our findings provide theoretical backing for Dynamic Computation Allocation. The core manifold evolution equation proves that breaking continuous semantics and executing rigorous logical clustering entails massive geometric distortion. This explains why complex logical reasoning (e.g., L4 primes) requires deeper network layers or extra computation (like Chain-of-Thought, CoT). Model processing is not a simple linear orthogonal projection, but a context-driven incremental logical clustering process: while maintaining basal topological preservation ($C_{ij}>0$) to prevent semantic collapse, it injects pure innovation components along specific dimensions to diverge. This specific divergence is critical for forging discrete boundaries. The resulting topological distortions in high-dimensional space are not residual decoupling noise; they are the geometric bedrock from which intelligent behavior emerges. This suggests future Transformer architectures could monitor the convergence gradient of $U_{sim}$ in early layers to dynamically trigger Early Exits or allocate extra computation, enabling adaptive reasoning. It also implies that understanding complex LLM reasoning must shift from searching for perfect linear subspaces to interrogating how models dynamically reshape manifold topologies across varying tasks.

Second, this research offers mechanistic insights into overcoming ``Sycophancy'' and hallucination \citep{huang_survey_2025}. The L5 experiment proves that when faced with non-logical directives, the model fails to pay the necessary algebraic cost to generate specific divergence ($U_{sim}<0$), leading to task feature separation failure. This implies that current scalar-reward alignment methods (e.g., RLHF or DPO) may harbor structural blind spots. Future alignment strategies could explicitly constrain the class-specific divergence ($U_{sim}$) between cross-class concepts within the loss function. Forcing the model to construct orthogonal divergence vectors for contradictory concepts during pre-training or fine-tuning could fundamentally enhance its robustness against misleading prompts.

\subsection{Limitations}

While our controlled experiments causally link topological deformation to logical classification, the methodological boundaries of this work must be rigorously defined. To obtain absolute algebraic baselines and crisp decision boundaries, we utilized numerical logic (magnitude, parity, primality) as the observational target. Although dual-modality experiments (Arabic and English) confirmed the high abstraction of this geometric transformation, multi-hop or commonsense reasoning in natural language involves far blurrier inter-class boundaries and multidimensional feature entanglement. Whether the single specific interference vector ($\Delta_{specific}$) extracted here scales losslessly to highly non-linear, composite semantic spaces requires validation on larger, more complex real-world corpora.

Furthermore, while our macro-geometric ``specific vector ablation'' successfully proved causality (via direct subtraction of $\Delta_{specific}$), it remains a coarse-grained algebraic intervention based on global feature vectors. Despite macro-level topological closure proving its high spatial specificity, we cannot yet precisely map how this algebraic erasure cascades to affect the internal circuits of specific Attention Heads or MLPs at the micro-scale. Additionally, we cannot entirely rule out whether forced algebraic subtraction in high-dimensional space introduces subtle non-linear artifacts in unobserved redundant subspaces. Future work must integrate Causal Mediation Analysis or Sparse Autoencoders (SAEs) to further untangle the precise mapping between macro-manifold evolution and underlying neuronal clusters.

Finally, the mathematical derivation of the ``equivalent rotation angle $\alpha$'' heavily relies on the equivalent hypersphere projection constraints afforded by RMSNorm. While Gram-Schmidt orthogonalization successfully isolated localized linear divergence vectors, the global non-linear dynamics of the forward pass may entail topological phase transitions more complex than the current dual-mechanism model. Future studies tracking cross-layer dynamics could dissect how specific divergence accumulates layer by layer. For early models lacking such normalization (or using absolute LayerNorm), the translation-to-rotation mapping formulas may require additional scaling corrections. However, we emphasize that the foundational tension between continuous topology and discrete logic, and the micro-antagonism between specific divergence and topological preservation, remain universal mechanisms independent of specific normalization techniques.

\subsection{Conclusion}

The representational geometry of Large Language Models has long been constrained by the theoretical presumption of ``Isometric Isomorphism,'' treating contextual modulation as smooth orthogonal translations. This study shatters that paradigm, precisely formalizing the dual algebraic effects of context-induced non-isometric manifold deformation.

Through Gram-Schmidt decomposition of residual streams and real-time causal specific vector ablation, we arrive at our ultimate conclusion: the emergence of rigorous logical behavior in LLMs is by no means the lossless retrieval of existing knowledge within a static space; rather, it is a violent, dynamic topological reshaping process in high-dimensional space. The model must leverage class-specific divergence to overcome basal topological preservation, paying an irreducible cost of geometric distortion to forcefully carve out discrete logical islands from a continuous semantic ocean. This discovery fills a mechanistic void regarding non-linear manifold dynamics within the field of Mechanistic Interpretability, providing fundamental principles and causal evidence essential for understanding and ultimately mastering the intelligent behaviors of Large Language Models.

\bibliographystyle{plainnat}
\bibliography{references}

@article{park_linear_2023,
	title = {The {Linear} {Representation} {Hypothesis} and the {Geometry} of {Large} {Language} {Models}},
	copyright = {Creative Commons Attribution Non Commercial Share Alike 4.0 International},
	url = {https://arxiv.org/abs/2311.03658},
	doi = {10.48550/ARXIV.2311.03658},
	urldate = {2026-03-23},
	journal = {arXiv},
	author = {Park, Kiho and Choe, Yo Joong and Veitch, Victor},
	year = {2023},
}

@article{hu_representational_2026,
	title = {The {Representational} {Geometry} of {Number}},
	copyright = {arXiv.org perpetual, non-exclusive license},
	url = {https://arxiv.org/abs/2602.06843},
	doi = {10.48550/ARXIV.2602.06843},
	urldate = {2026-03-23},
	journal = {arXiv},
	author = {Hu, Zhimin and Niu, Lanhao and Varma, Sashank},
	year = {2026},
}

@inproceedings{yang_unifying_2025,
	title={Unifying Attention Heads and Task Vectors via Hidden State Geometry in In-Context Learning},
    author={Haolin Yang and Hakaze Cho and Yiqiao Zhong and Naoya Inoue},
    booktitle={The Thirty-ninth Annual Conference on Neural Information Processing Systems},
    year={2025},
    url={https://openreview.net/forum?id=FIfjDqjV0B}
}

@inproceedings{zhou_geometry_2025,
    title={The Geometry of Reasoning: Flowing Logics in Representation Space},
    author={Yufa Zhou and Yixiao Wang and Xunjian Yin and Shuyan Zhou and Anru Zhang},
    booktitle={The Fourteenth International Conference on Learning Representations},
    year={2026},
    url={https://openreview.net/forum?id=ixr5Pcabq7}
}

@article{zhang_root_2019,
	title = {Root {Mean} {Square} {Layer} {Normalization}},
	copyright = {arXiv.org perpetual, non-exclusive license},
	url = {https://arxiv.org/abs/1910.07467},
	doi = {10.48550/ARXIV.1910.07467},
	urldate = {2026-03-23},
	journal = {arXiv},
	author = {Zhang, Biao and Sennrich, Rico},
	year = {2019},
}

@inproceedings{jin_exploring_2025,
	title = "Exploring Concept Depth: How Large Language Models Acquire Knowledge and Concept at Different Layers?",
    author = "Jin, Mingyu  and
      Yu, Qinkai  and
      Huang, Jingyuan  and
      Zeng, Qingcheng  and
      Wang, Zhenting  and
      Hua, Wenyue  and
      Zhao, Haiyan  and
      Mei, Kai  and
      Meng, Yanda  and
      Ding, Kaize  and
      Yang, Fan  and
      Du, Mengnan  and
      Zhang, Yongfeng",
    editor = "Rambow, Owen  and
      Wanner, Leo  and
      Apidianaki, Marianna  and
      Al-Khalifa, Hend  and
      Eugenio, Barbara Di  and
      Schockaert, Steven",
    booktitle = "Proceedings of the 31st International Conference on Computational Linguistics",
    month = jan,
    year = "2025",
    address = "Abu Dhabi, UAE",
    publisher = "Association for Computational Linguistics",
    url = "https://aclanthology.org/2025.coling-main.37/",
    pages = "558--573"
}

@article{xu_low-dimensional_2026,
	title = {Low-{Dimensional} {Execution} {Manifolds} in {Transformer} {Learning} {Dynamics}: {Evidence} from {Modular} {Arithmetic} {Tasks}},
	copyright = {Creative Commons Attribution 4.0 International},
	shorttitle = {Low-{Dimensional} {Execution} {Manifolds} in {Transformer} {Learning} {Dynamics}},
	url = {https://arxiv.org/abs/2602.10496},
	doi = {10.48550/ARXIV.2602.10496},
	urldate = {2026-03-23},
	journal = {arXiv},
	author = {Xu, Yongzhong},
	year = {2026},
}

@inproceedings{hindupur_projecting_nodate,
    title={Projecting Assumptions: The Duality Between Sparse Autoencoders and Concept Geometry},
    author={Sai Sumedh R. Hindupur and Ekdeep Singh Lubana and Thomas Fel and Demba E. Ba},
    booktitle={ICML 2025 Workshop on Methods and Opportunities at Small Scale},
    year={2025},
    url={https://openreview.net/forum?id=AKaoBzhIIF}
}

@inproceedings{zhou_lssf_2025,
	address = {Vienna, Austria},
	title = {{LSSF}: {Safety} {Alignment} for {Large} {Language} {Models} through {Low}-{Rank} {Safety} {Subspace} {Fusion}},
	isbn = {979-8-89176-251-0},
	shorttitle = {{LSSF}},
	url = {https://aclanthology.org/2025.acl-long.1479/},
	doi = {10.18653/v1/2025.acl-long.1479},
	urldate = {2026-03-23},
	booktitle = {Proceedings of the 63rd {Annual} {Meeting} of the {Association} for {Computational} {Linguistics} ({Volume} 1: {Long} {Papers})},
	publisher = {Association for Computational Linguistics},
	author = {Zhou, Guanghao and Qiu, Panjia and Chen, Cen and Li, Hongyu and Chu, Jason and Zhang, Xin and Zhou, Jun},
	editor = {Che, Wanxiang and Nabende, Joyce and Shutova, Ekaterina and Pilehvar, Mohammad Taher},
	month = jul,
	year = {2025},
	pages = {30621--30638},
}

@article{huang_survey_2025,
	title = {A {Survey} on {Hallucination} in {Large} {Language} {Models}: {Principles}, {Taxonomy}, {Challenges}, and {Open} {Questions}},
	volume = {43},
	issn = {1046-8188, 1558-2868},
	shorttitle = {A {Survey} on {Hallucination} in {Large} {Language} {Models}},
	url = {https://dl.acm.org/doi/10.1145/3703155},
	doi = {10.1145/3703155},
	language = {en},
	number = {2},
	urldate = {2026-03-23},
	journal = {ACM Transactions on Information Systems},
	author = {Huang, Lei and Yu, Weijiang and Ma, Weitao and Zhong, Weihong and Feng, Zhangyin and Wang, Haotian and Chen, Qianglong and Peng, Weihua and Feng, Xiaocheng and Qin, Bing and Liu, Ting},
	month = mar,
	year = {2025},
	pages = {1--55},
}

@article{zhang_human-like_2026,
	title = {Human-like {Social} {Compliance} in {Large} {Language} {Models}: {Unifying} {Sycophancy} and {Conformity} through {Signal} {Competition} {Dynamics}},
	copyright = {arXiv.org perpetual, non-exclusive license},
	shorttitle = {Human-like {Social} {Compliance} in {Large} {Language} {Models}},
	url = {https://arxiv.org/abs/2601.11563},
	doi = {10.48550/ARXIV.2601.11563},
	urldate = {2026-03-23},
	journal = {arXiv},
	author = {Zhang, Long and Chen, Wei-neng},
	year = {2026},
}

@article{boot_eye_2022,
	title = {An eye tracking experiment investigating synonymy in conceptual model validation},
	volume = {47},
	issn = {14670895},
	url = {https://linkinghub.elsevier.com/retrieve/pii/S1467089522000306},
	doi = {10.1016/j.accinf.2022.100578},
	language = {en},
	urldate = {2026-03-23},
	journal = {International Journal of Accounting Information Systems},
	author = {Boot, Walter R. and Dunn, Cheryl L. and Fulmer, Bachman P. and Gerard, Gregory J. and Grabski, Severin V.},
	month = dec,
	year = {2022},
	pages = {100578},
}

@article{joshi_geometry_2025,
	title = {Geometry of {Decision} {Making} in {Language} {Models}},
	copyright = {Creative Commons Attribution Share Alike 4.0 International},
	url = {https://arxiv.org/abs/2511.20315},
	doi = {10.48550/ARXIV.2511.20315},
	urldate = {2026-03-23},
	journal = {arXiv},
	author = {Joshi, Abhinav and Bhatt, Divyanshu and Modi, Ashutosh},
	year = {2025},
}

\end{document}